\documentclass[final,5p,times,twocolumn]{elsarticle}

\usepackage{graphicx}
\usepackage{algorithm}
\usepackage{algorithmic}
\usepackage{array}
\usepackage{amsmath}

\journal{ArXiv}

\begin{document}
\begin{frontmatter}
\title{End-to-end Adversarial Learning for Generative Conversational Agents}
\author{Oswaldo Ludwig}

\begin{abstract}
This paper presents a new adversarial learning method for generative conversational agents (GCA) besides a new model of GCA. Similar to previous works on adversarial learning for dialogue generation, our method assumes the GCA as a generator that aims at fooling a discriminator that labels dialogues as human-generated or machine-generated; however, in our approach, the discriminator performs token-level classification, i.e. it indicates whether the current token was generated by humans or machines. To do so, the discriminator also receives the context utterances (the dialogue history) and the incomplete answer up to the current token as input. This new approach makes possible the end-to-end training by backpropagation. Moreover, the trained discriminator can be used to choose the best answer among the answers generated by different trained models to improve the performance even more. A self-conversation process enables to produce a set of generated data with more diversity for the adversarial training. This approach improves the performance on questions not related to the training data. Experimental results with human and adversarial evaluations show that the adversarial method yields significant performance gains over the usual teacher forcing training.
\end{abstract}
\end{frontmatter}

\section{Introduction}
\label{intro}

Neural language modeling \cite{bengio2003neural}, \cite{mikolov2010recurrent} uses recurrent neural networks to create effective models for different tasks in Natural Language Processing, such as open domain dialogue generation \cite{ritter2011data}, \cite{serban2017hierarchical}, which aims at generating meaningful and coherent dialogue responses given the dialogue history. This task is the subject of this paper.

Conversational agents can be divided into two main classes: retrieval-based and generative agents. The retrieval-based model does not generate new text, it access a repository of predefined responses and chooses an appropriate response based on the input. This paper is about generative models, which generate new responses from the scratch, usually based on the sequence to sequence modeling \cite{vinyals2015neural}. In this context, we present a new end-to-end adversarial learning method for generative conversational agents (GCA) besides a new model of GCA.

The paper is organized as follows: Section \ref{bib} briefly reports the state-of-the-art in open domain dialogue generation, while Section \ref{model} presents the proposed model of GCA. In Section \ref{training} we explain our new end-to-end adversarial training. Section \ref{sec_experiments} reports the experiments, while Section \ref{conclusion} summarizes some conclusions.

\section{State-of-the-art}
\label{bib}

Sequence to sequence (seq2seq) modeling is being successfully applied to neural machine translation \cite{kalchbrenner2013recurrent}, \cite{sutskever2014sequence}, since this model is language-independent and able to implicitly learn semantic \cite{ludwig2017learning}, syntactic and contextual dependencies \cite{sordoni2015neural}. Further advances in end-to-end training with this model has made it possible to build successful systems \cite{johnson2016google} for different natural language tasks, including parsing \cite{vinyals2015grammar}, image captioning \cite{vinyals2015show}, and open domain dialogue generation \cite{vinyals2015neural}.

The research on GCA poses several challenges, such as the sensitivity to local and global context. GCA models tend to generate safe responses regardless of the input, such as  ``I am not sure''. In \cite{li2015diversity} the authors propose to replace the usual maximum likelihood training objective functions, which favor generic responses of higher frequency in the training data sets, by a generalization of the Maximum Mutual Information (MMI), to avoid favoring responses that unconditionally present high probability. Other approaches build on this MMI-based strategy, such as seq2BF \cite{mou2016sequence}, which proposes a backward and forward sequences model that generates a reply built around a keyword candidate belonging to the context. This keyword is chosen by maximizing the point-wise mutual information (PMI) on the context sentence. The sensitivity to local context was also approached by transfer learning, such as in the topic augmented neural response generation, presented in \cite{xiong2016neural}, where a pre-trained probabilistic topic model provides information on the context to the decoder together with the thought vector provided by the encoder. Our GCA model presents the encoded context (i.e. the thought vector) to the decoder at each decoding iteration, rather than using it only to set up the initial state of the decoder, to avoid favoring short and unconditional responses with high prior probability.

Regarding the long-term dependencies, i.e. the global context sensitivity, we highlight the work \cite{sordoni2015hierarchical}, which proposes a hierarchical recurrent encoder-decoder model (HRED) that models a user search session as two hierarchical sequences: a sequence of sentences and a sequence of words in each query. This problem was also approached by \cite{serban2016building}, which proposed a model that consists of three RNN modules: an encoder RNN, a context RNN, and a decoder RNN. A sequence of tokens is encoded into a real-valued thought vector by the encoder RNN. The sequence of thought vectors is given as input to the context RNN, which updates its internal hidden state retaining the contextual information up to that point in time into a real-valued context vector, which conditions the decoder RNN. We intend to approach long-term dependencies by using hierarchical LSTMs in the encoder of our model as future work.

Another open issue in GCA is the speaker consistency. The GCA response may simulate different persons belonging to the training data, yielding inconsistent responses. An interesting work on this issue is \cite{li2016persona} that shows that seq2seq models provide a straightforward mechanism for incorporating persona as embeddings. This work proposes a persona embedding that permits the incorporation of background facts for user profiles, person-specific language behavior, and interaction style. The modeling style of our GCA makes it easy to incorporate persona as embeddings.

Our work is closely related to the work of Li et al. \cite{li2017adversarial}, in which the authors borrow the idea of adversarial training \cite{goodfellow2014generative} to generate utterances indistinguishable from human utterances. The model is composed of a neural seq2seq model, which defines the probability of generating a dialogue sequence, and a discriminator that labels dialogue utterances as human-generated or machine-generated, similar to the evaluator in the Turing test. The authors cast the task as a reinforcement learning problem. Our work follows this research line, but with different approaches, such as the end-to-end training by backpropagation, as explained in Section \ref{training}.

\section{The new model of generative conversational agent}
\label{model}

In this section, we define the proposed GCA and show the relationship between this model and the canonical seq2seq model.

\begin{figure}[ht]
\vskip 0.0in
\begin{center}
\centerline{\includegraphics[width=\columnwidth]{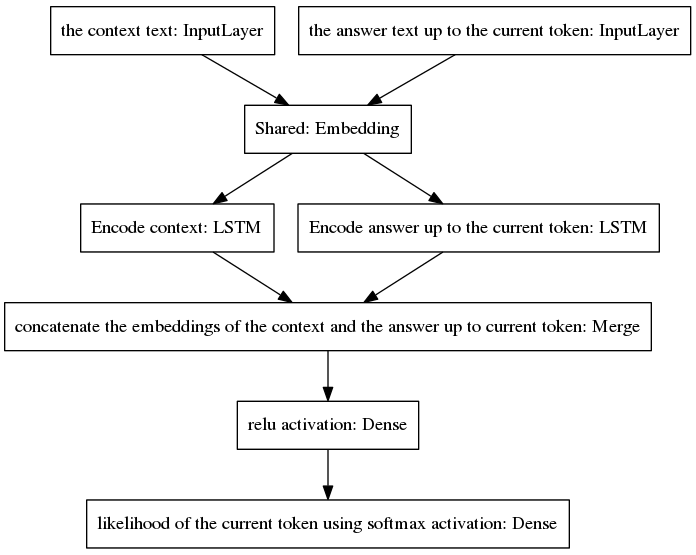}}
\caption{The model of the new GCA.}
\label{figModel}
\end{center}
\vskip -0.3in
\end{figure}

The canonical seq2seq model became popular in neural machine translation, a task that has different prior probability distributions for the words belonging to the input and output sequences, since the input and output utterances are written in different languages. The GCA architecture presented here assumes the same prior distributions for input and output words. Therefore, it shares an embedding layer (Glove pre-trained word embedding\footnote{https://nlp.stanford.edu/projects/glove/}) between the encoding and decoding processes through the adoption of a new model. To improve the sensitivity to the context, the thought vector (i.e. the encoder output) encodes the last $N_u$ utterances of the conversation up to the current point (the dialogue history). The thought vector is concatenated with a dense vector that encodes the incomplete answer generated up to the current point, to avoid forgetting the context during the decoding process, as will be explained in Section \ref{diff}. The resulting vector is provided to dense layers that predict the current token of the answer, as can be seen in Figure \ref{figModel}. Therefore, our model is different from the canonical model, in which the encoder output is used only to set up the initial state of the decoder. 

The dialogue history/context utterances are arranged as a vector $\textbf{x} \in \mathbb{R}^{s_s}$ containing a sequence of token indexes that is padded with zeros to have dimension $s_s$, i.e. an arbitrary value for the sentence length. The elements $x_i$, $i \in \left\{1\ldots s_s\right\}$, of $\textbf{x}$ are encoded into one-hot vector representation $\bar{x}_i \in \mathbb{R}^{s_v}$, where $s_v$ is the size of the adopted vocabulary. The same happens with the elements $y_i$ from the incomplete answer $\textbf{y}\in \mathbb{R}^{s_s}$. These vectors are arranged to compose the matrices $X=\left[\bar{x}_1\left.\right. \bar{x}_2\left.\right.\ldots \bar{x}_{s_s}\right]$ and $Y=\left[\bar{y}_1\left.\right. \bar{y}_2\left.\right.\ldots \bar{y}_{s_s}\right]$. These matrices are processed by the embedding layer, represented by the matrix $W_e \in \mathbb{R}^{s_e \times s_v}$, where $s_e$ is the arbitrary dimension of the word embedding vector, yielding two dense matrices $E_c \in \mathbb{R}^{s_e \times s_s}$ and $E_a \in \mathbb{R}^{s_e \times s_s}$:  
\begin{equation}
\label{model0}
\begin{array}{l}
			E_c= W_e X \\ 
			E_a= W_e Y \\ 
\end{array}
\end{equation}

The proposed model uses two Long Short-term Memory networks (LSTM) \cite{hochreiter1998vanishing}, both with the same architecture, one to process $E_c$, which is related to the dialogue history/context, and another to process $E_a$, which is related to the incomplete answer generated up to the current point/iteration $t$. Details on the LSTM mathematical model can be found in \cite{hochreiter1998vanishing}; therefore, we will represent the LSTMs simply as the functions $\Gamma_c: \mathbb{R}^{s_e \times s_s}\rightarrow \mathbb{R}^{s_{se}}$ and $\Gamma_a: \mathbb{R}^{s_e \times s_s}\rightarrow \mathbb{R}^{s_{se}}$, where $s_{se}$ is the arbitrary dimension of the sentence embedding\footnote{The implementation of the LSTM algorithm receives as input a 3D tensor whose shape also includes the batch size.}. These functions extract the sentence embedding vectors of the dialogue history and the incomplete answer
\begin{equation}
\label{model1}
\begin{array}{l}
			e_c = \Gamma_c\left(E_c;\mathcal{W}_c\right) \\ 
			e_a = \Gamma_a\left(E_a;\mathcal{W}_a\right) \\ 
\end{array}
\end{equation}
respectively. $\mathcal{W}_c$ and $\mathcal{W}_a$ are sets of parameters of the LSTMs. These vectors are concatenated and provided to dense layers that output the vector $\textbf{p} \in \mathbb{R}^{s_v}$ encoding the probability $p\left(v_j  \middle| \textbf{x},\textbf{y}\right)$ of each token $v_j$, $j \in \left\{1 \ldots s_v\right\}$, of the adopted vocabulary. The dense layers are modeled as follows:
\begin{equation}
\label{model2}
\begin{array}{l}
			e = \left[e_c\left.\right. e_a\right] \\
			y_h=\sigma\left(W_{1\left. \right.} e+b_1\right) \\ 
      \textbf{p}=\varphi\left(W_{2\left. \right.} y_h+b_2\right) \\       
\end{array}
\end{equation}
where $W_1$ and $W_2$ are matrices of synaptic weights,  $b_1$ and $b_2$ are bias vectors, $\sigma\left(\cdot\right)$ is the relu activation function, and $\varphi\left(\cdot\right)$ is the softmax activation function. 

\begin{flushleft}
\end{flushleft}

We adopt the greedy decoding; therefore, the algorithm iterates by making the position $t + 1$ of the incomplete answer $\textbf{y}$ equal to the index of the largest element of $\textbf{p}$, and feeding it back to the input layer on the right-hand side of the model shown in Figure \ref{figModel}. This process continues until the token representing the end of the sentence is predicted, as can be seen in the simplified Algorithm \ref{greedy}. The training follows the teacher forcing strategy. The software package of this GCA is available on GitHub \cite{oswaldo_ludwig_2017_825303}.

\begin{algorithm}[ht!]
\suppressfloats[t]
 \caption{New GCA with greedy decoding}
\footnotesize 
 \label{greedy}
 \begin{algorithmic}[1]
    \STATE {\textbf{Input:} $\textbf{x}$: the input sequence (context text).}\\
    \STATE {\textbf{Output:} $\textbf{y}$, $p$: the sampled output sequence and its conditional probability $p(\textbf{y}|\textbf{x})$.}\\
		\STATE {$p \leftarrow 1$}\\
		\STATE {$ \textbf{y} \leftarrow []$}\\
    \STATE {$y \leftarrow$ 'BOS' (symbol representing the beginning of the sentence);}\\
    \WHILE {$y <>$ 'EOS' (symbol representing the end of sentence)}
			\STATE {$\textbf{y} \leftarrow [\textbf{y}, y]$;}\\
			\STATE {input $\textbf{x}$ and $\textbf{y}$ into the two input layers of the model (\ref{model0}) - (\ref{model2}) (see Figure \ref{figModel});}\\
			\STATE {$y \leftarrow$ token corresponding to the largest output of the model (\ref{model0}) - (\ref{model2});}\\
			\STATE {$p\left(y\middle|\textbf{x}, \textbf{y}\right) \leftarrow$ the value of the larger output of the model (\ref{model0}) - (\ref{model2});}\\
			\STATE {$p \leftarrow p \times p\left(y\middle|\textbf{x}, \textbf{y}\right)$;}\\
		\ENDWHILE
    \end{algorithmic}
\end{algorithm}

\subsection{Differences between our model and the canonical seq2seq}
\label{diff}

Short responses with high prior probability, regardless of the input, is a common output of the canonical seq2seq model; however, this behavior is not observed in our model. We believe this is due to its architecture. Seq2seq modeling approximates $\textit{p}\left(\textbf{y}\middle|\textbf{x}\right)$ by $\textbf{\textit{p}}_\theta\left(\textbf{y}\middle|g\left(\textbf{x}\right)\right)$, where $g\left(\textbf{x}\right)$ denotes the encoder output, i.e. the thought vector. Then, the decoder generates answers using this learned model. In the case of the canonical seq2seq this model can be expressed as:
\begin{equation}
\label{canonical1}
\textbf{\textit{p}}_\theta\left(\textbf{y}\middle|g\left(\textbf{x}\right)\right) = \prod_{i=1}^{s_s} p_\theta\left(y_i \middle| h_{i-1}, y_{i-1} \right)
\end{equation}
where $y_0$ is the index of the token representing the beginning of the sentence and $g\left(\textbf{x}\right)$ is used only to set up the initial state $h_0$ of the decoder, which is updated along the decoding iterations $i$ as follows:
\begin{equation}
\label{canonical2}
\begin{array}{l}
			h_0 = g\left(\textbf{x}\right) \\
			h_1 = f_\alpha\left(g\left(\textbf{x}\right),y_0\right) \\
			h_2 = f_\alpha\left(f_\alpha\left(g\left(\textbf{x}\right),y_0\right),y_1\right) \\
			\vdots \\
      h_{s_s-1} = f_\alpha \left(\ldots f_\alpha\left(f_\alpha\left(g\left(\textbf{x}\right),y_0\right),y_1\right) \ldots y_{s_s - 2}\right) \\      
\end{array}
\end{equation}
where $f_\alpha$ represents the set of operations that the input and forget gates apply on the state variables and $\alpha \in \theta$ is the set of parameters of these gates, assuming an LSTM as the decoder. Notice that the nested application of operations on $\textbf{x}$, such as the operation applied by the forget gate\footnote{Note that the outputs of the forget gate are within the interval $[0, 1]$ due to its sigmoid activation function. Thus, the influence of $g\left(\textbf{x}\right)$ may decrease exponentially over the decoding iterations.}, can erase information about the context along the decoder iterations, resulting in the usual safe answers, regardless of the input $\textbf{x}$. On the other hand, our GCA architecture models $\textit{p}\left(\textbf{y}\middle|\textbf{x}\right)$ in another way:
\begin{equation}
\label{canonical3}
\textbf{\textit{p}}_\theta\left(\textbf{y}\middle|g\left(\textbf{x}\right)\right) = \prod_{i=1}^{s_s} p_\theta\left(y_i \middle| f_\beta\left(y_0 \ldots y_{i-1}\right), g\left(\textbf{x}\right) \right)
\end{equation}
where $f_\beta\left(\cdot\right)$ (with $\beta \in \theta$) represents the LSTM that encodes the incomplete answer $(y_0 \ldots y_{i-1})$. Since the encoder output $g\left(\textbf{x}\right)$ is provided to the decoder at each decoding iteration $i$, it is not subject to the nested functions of (\ref{canonical2}).

Another important feature of our model is the use of the distributional syntactic and semantic information encoded into the word embedding also in the decoding process, as can be seen in the second equation of (\ref{model0}).

\section{End-to-end Adversarial Training by Backpropagation}
\label{training}

Similar to the work of Li et al. \cite{li2017adversarial}, our adversarial training assumes a data set $\mathcal{H}$ of human-generated dialogue utterances, a generator $G$ and a discriminator $D$. Our work also assumes the GCA as the generator $G$ that learns to fool the discriminator $D$; however, in our approach the discriminator $D$ performs token-level classification, rather than sentence-level classification, i.e. $D$ is a binary classifier that outputs a label indicating whether the current token was generated by humans or machines. To do so, the discriminator takes as input the current token (denoted $y^-$, if it is generated by $G$, and $y^+$, if it comes from a utterance selected from $\mathcal{H}$), the previous $N_u$ dialogue utterances (denoted $\textbf{x}^-$, if it is generated by $G$, and $\textbf{x}^+$, if it is from $\mathcal{H}$), and a sequence representing the incomplete answer (denoted $\textbf{y}^-$, if it is generated by $G$, and $\textbf{y}^+$, if it is part of an utterance from $\mathcal{H}$). 

\begin{figure}[ht]
\vskip 0.0in
\begin{center}
\centerline{\includegraphics[width=\columnwidth]{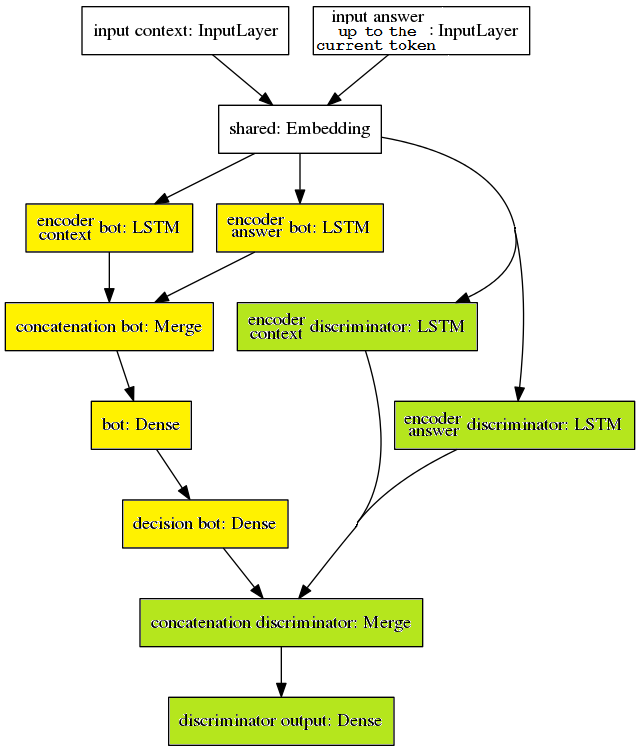}}
\caption{Model composed by the generator and the discriminator for the end-to-end adversarial learning. The yellow blocks belong to the GCA (the generator), while the green blocks belong to the discriminator. The white blocks are shared between generator and discriminator.}
\label{GAN}
\end{center}
\vskip -0.3in
\end{figure}

The inputs of $D$ are processed by (\ref{model0}) yielding the matrices $E_c$ and $E_a$ that are processed by two LSTMs, $\Gamma_{cd}: \mathbb{R}^{s_e \times s_s}\rightarrow \mathbb{R}^{s_{sed}}$ and $\Gamma_{ad}: \mathbb{R}^{s_e \times s_s}\rightarrow \mathbb{R}^{s_{sed}}$, where $s_{sed}$ is the arbitrary dimension of the sentence embedding vectors of $D$. $\Gamma_{cd}$ encodes the $N_u$ previous utterances (i.e. the context) and $\Gamma_{ad}$ encodes the incomplete answer up to the current token, yielding two sentence embedding vectors:
\begin{equation}
\label{model3}
\begin{array}{l}
			e_{cd} = \Gamma_{cd}\left(E_c;\mathcal{W}_{cd}\right) \\ 
			e_{ad} = \Gamma_{ad}\left(E_a;\mathcal{W}_{ad}\right) \\ 
\end{array}
\end{equation}
where $\mathcal{W}_{cd}$ and $\mathcal{W}_{ad}$ are sets of parameters of the LSTMs of $D$. These vectors are concatenated with the generator output $\textbf{p}$ and provided to a dense layer with sigmoid activation function that outputs $l \in \left[0,1\right]$, with 1 corresponding to a perfect match with the class \textit{human-generated} and 0 to the class \textit{machine-generated}. The dense layer is modeled as follows:
\begin{equation}
\label{model4}
\begin{array}{l}
			e_d = \left[\textbf{p}\left.\right. e_{cd}\left.\right. e_{ad}\right] \\
			l=\alpha\left(W_{d\left. \right.} e_d+b_d\right) \\        
\end{array}
\end{equation}
where $W_d$ is the matrix of synaptic weights,  $b_d$ is the bias vector, and $\alpha\left(\cdot\right)$ is the sigmoid activation function. This new approach makes possible end-to-end training by backpropagation using the model of Figure \ref{GAN}. 
\begin{flushleft}
\end{flushleft}

The white blocks of Figure \ref{GAN} are shared between generator and discriminator and are modeled by (\ref{model0}), the yellow blocks, whose model is given by (\ref{model1}) and (\ref{model2}), compose $G$, while the green blocks, whose model is given by (\ref{model3}) and (\ref{model4}), compose $D$.

The adversarial training starts by using a version of the GCA\footnote{The model given by (\ref{model0})-(\ref{model2}) that corresponds to Figure \ref{figModel}} pre-trained by teacher forcing to generate a set $\mathcal{M}$ containing pairs of machine-generated dialogue utterances.

Our generation process is different from Li et al. \cite{li2017adversarial}, in which the machine-generated dialogue is the GCA answer to utterances belonging to $\mathcal{H}$, as can be seen in Figure 1 of \cite{li2017adversarial}. Namely, rather than produce the pair $\left(\textbf{x}^+, \textbf{y}^-\sim \mathcal{G}\left(\cdot\left|\right.\textbf{x}^+\right)\right)$, where $\mathcal{G}$ represents the probability distribution modeled by $G$, our algorithm produces a more diverse machine-generated data set by iterating over utterances generated by a self-conversation process. The algorithm selects a utterance $\textbf{x}^+$ from the training data set and uses it as a seed that is provided to $G$, then it iterates twice\footnote{The number of iteration can be arbitrary.} by feeding back its own output, in such a way to compose the set $\mathcal{M} = \left\{\left(\textbf{x}^-_i\sim \mathcal{G}\left(\cdot\left|\right.\textbf{x}^+_i\right), \textbf{y}^-_i\sim \mathcal{G}\left(\cdot\left|\right.\textbf{x}^-_i\right)\right)\right\}_{i=1}^{N_m}$. In other words, the seed $\textbf{x}^+$ yields $\textbf{x}^-$, which is in turn feed back to $G$ yielding $\textbf{y}^-$.  This approach improves the performance on unseen data, since $G$ is trained to produce answers indistinguishable from human answers for context utterances that are not present in $\mathcal{H}$.

The adversarial training follows by training only the discriminator model $D$ (i.e. (\ref{model0}), (\ref{model3}), and (\ref{model4})) for $N_D$ epochs using data from $\mathcal{H}$ and $\mathcal{M}$, along with their labels: 1 for examples from $\mathcal{H}$ and 0 for examples from $\mathcal{M}$. Then the trained weights of $D$ (i.e. $\mathcal{W}_{cd}$, $\mathcal{W}_{ad}$, $W_d$, and $b_d$) are imported to the green blocks of the model of Figure \ref{GAN}, frozen, and the weights of $G$ (i.e. $\mathcal{W}_{c}$, $\mathcal{W}_{a}$, $W_1$, $W_2$, $b_1$, and $b_2$) are updated by back-propagation for $N_G$ epochs using only data from $\mathcal{M}$ along with the target labels, which are all equal to one, since the idea is to train $G$ (the yellow blocks) to fool $D$ (the green blocks), in such a way to induce $D$ to output 1 (the label of human-generated tokens) for the tokens generated by $G$.

The adversarial training is interleaved with teacher forcing using $\mathcal{H}$ as training data, since the idea is to learn to generate human-like answers for context utterances from $\mathcal{M}$ without forgetting the proper answers to utterances from $\mathcal{H}$. After training $G$, its weights $\mathcal{W}_{c}$, $\mathcal{W}_{a}$, $W_1$, $W_2$, $b_1$, and $b_2$ are imported to the GCA (Figure \ref{figModel}) to generate a new set $\mathcal{M}$ and start a new iteration, running again all the previous steps, as summarized in Algorithm \ref{training_alg}.

\begin{algorithm}[ht!]
\suppressfloats[t]
 \caption{End-to-end adversarial training}
\footnotesize 
 \label{training_alg}
 \begin{algorithmic}[1]
    \STATE {\textbf{Input:} $\mathcal{H}$  (human-generated data set).}\\
    \STATE {\textbf{Output:} $\mathcal{W}_{cd}$, $\mathcal{W}_{ad}$, $W_d$, $b_d$, $\mathcal{W}_{c}$, $\mathcal{W}_{a}$, $W_1$, $W_2$, $b_1$, and $b_2$  (parameters of $D$ and $G$).}\\
		\STATE {train the GCA of Figure \ref{figModel} ((\ref{model0})-(\ref{model2})) by teacher forcing on $\mathcal{H}$ (using categorical cross-entropy loss).}
		\FOR {number of adversarial training epochs}
		\STATE {use the GCA to generate $\mathcal{M} = \left\{\left(\textbf{x}^-_i\sim \mathcal{G}\left(\cdot\left|\right.\textbf{x}^+_i\right), \textbf{y}^-_i\sim \mathcal{G}\left(\cdot\left|\right.\textbf{x}^-_i\right)\right)\right\}_{i=1}^{N_m}$.}
		\STATE {update $D$ (i.e. (\ref{model0}), (\ref{model3}), and (\ref{model4})) for $N_D$ epochs using $\mathcal{H}$ and $\mathcal{M}$, along with their labels: 1 for examples from $\mathcal{H}$ and 0 for examples from $\mathcal{M}$ (using binary cross-entropy loss).} 
		\STATE {import the updated weights $\mathcal{W}_{cd}$, $\mathcal{W}_{ad}$, $W_d$, and $b_d$ from $D$ to the green blocks of the model of Figure \ref{GAN}.}
		\STATE {import the weights $\mathcal{W}_{c}$, $\mathcal{W}_{a}$, $W_1$, $W_2$, $b_1$, and $b_2$ from the GCA to the yellow blocks of the model of Figure \ref{GAN}.}
		\STATE {freeze the weights of $D$ in the model of Figure \ref{GAN} and update the weights of $G$ (i.e. $\mathcal{W}_{c}$, $\mathcal{W}_{a}$, $W_1$, $W_2$, $b_1$, and $b_2$) by back-propagation for $N_G$ epochs using only $\mathcal{M}$ along with the target labels, which are all equal to one (using MSE loss function).}
		\STATE {import the updated weights $\mathcal{W}_{c}$, $\mathcal{W}_{a}$, $W_1$, $W_2$, $b_1$, and $b_2$ from $G$ to the GCA of Figure \ref{figModel}.}
		\STATE {update the GCA ((\ref{model0})-(\ref{model2})) by teacher forcing for $N_{tf}$ epochs on $\mathcal{H}$ (using categorical cross-entropy loss).}
		\ENDFOR
     \end{algorithmic}
\end{algorithm}

Since our discriminator performs token-level classification, we compose its outputs to evaluate the probability of $\textbf{y}=\textbf{y}^+$, i.e. to check whether the utterance is a human-generated answer. To do so, we provide the dialogue history/context $\textbf{x}$ to $D$. Assuming that the discriminator output at the $i^{th}$ iteration, $l_i \in \left[0,1\right]$, informs the probability $p\left(y^+_i \middle| y_0 \ldots y_{i-1}, \textbf{x}\right)$, i.e. the probability of $y_i = y^+_i$, it is possible to apply the chain rule as follows:
\begin{equation}
\label{prob}
p\left(\textbf{y}^+ \middle| \textbf{x}\right) = \prod_{i=1}^{s_s^*} p\left(y_i^+ \Bigg| \bigcap_{j=1}^{i-1} y_j ,\left.\right. \textbf{x}\right) = \prod_{i=1}^{s_s^*} l_i
\end{equation}
where $s_s^*$ is the effective number of tokens of the sentence, not the arbitrary length of the sentence after the padding $s_s$.   

The trained discriminator is also used to select the best answer among the answers generated by two different models, one trained by teacher forcing and another trained by our end-to-end adversarial method.

\section {Experiments}
\label{sec_experiments}

Evaluating open domain dialogue generation requires a human level of comprehension and background knowledge \cite{ludwig2016deepemb}. There are different quality attributes to be analyzed, such as the accuracy of the text synthesis \cite{mctear2016evaluating}, the ability to convey personality \cite{li2016persona}, the ability to maintain themed discussion \cite{xiong2016neural}, the ability to respond to specific questions, the ability to read and respond to moods of human participants \cite{zhou2017emotional}, the ability to show awareness of trends and social context \cite{applin2015new}, and the ability to detect intent \cite{marschner2014identification}.

Liu at al. \cite{liu2016not} shown that metrics from machine translation and automatic summarization, such as  BLEU \cite{papineni2002bleu}, METEOR \cite{lavie2007meteor} and ROUGE \cite{lin2004rouge} present either weak or no correlation with human judgments. Therefore, in this section we use human and adversarial evaluation to compare our new training method with the usual teacher forcing. 

Our human-generated data set $\mathcal{H}$ is composed of open domain dialogue utterances collected from English courses online. We chose this data source due to its linguistic accuracy, the variety of themes, and didactic format, i.e. if it is good for human learning, it should be good for the machines too.

We trained two GCA models using $\mathcal{H}$, one by teacher forcing and another by the new adversarial method, both models with the same architecture.

The models were implemented in Keras with Theano backend and trained using Adam optimizer. The adopted parameters were: $N_u=2$, $N_G=1$, $N_D=15$, $N_{tf}=1$, $N_m=7900$, $s_{e}=100$, $s_{se}=300$, $s_{sed}=300$, $s_{v}=7000$, $s_{s}=50$, learning rate of the generator $\alpha_g=5e^{-5}$, and learning rate of the discriminator $\alpha_d=1e^{-4}$. The adopted loss functions can be seen in Algorithm \ref{training_alg}.

\begin{figure}[ht]
\vskip 0.0in
\begin{center}
\centerline{\includegraphics[width=\columnwidth]{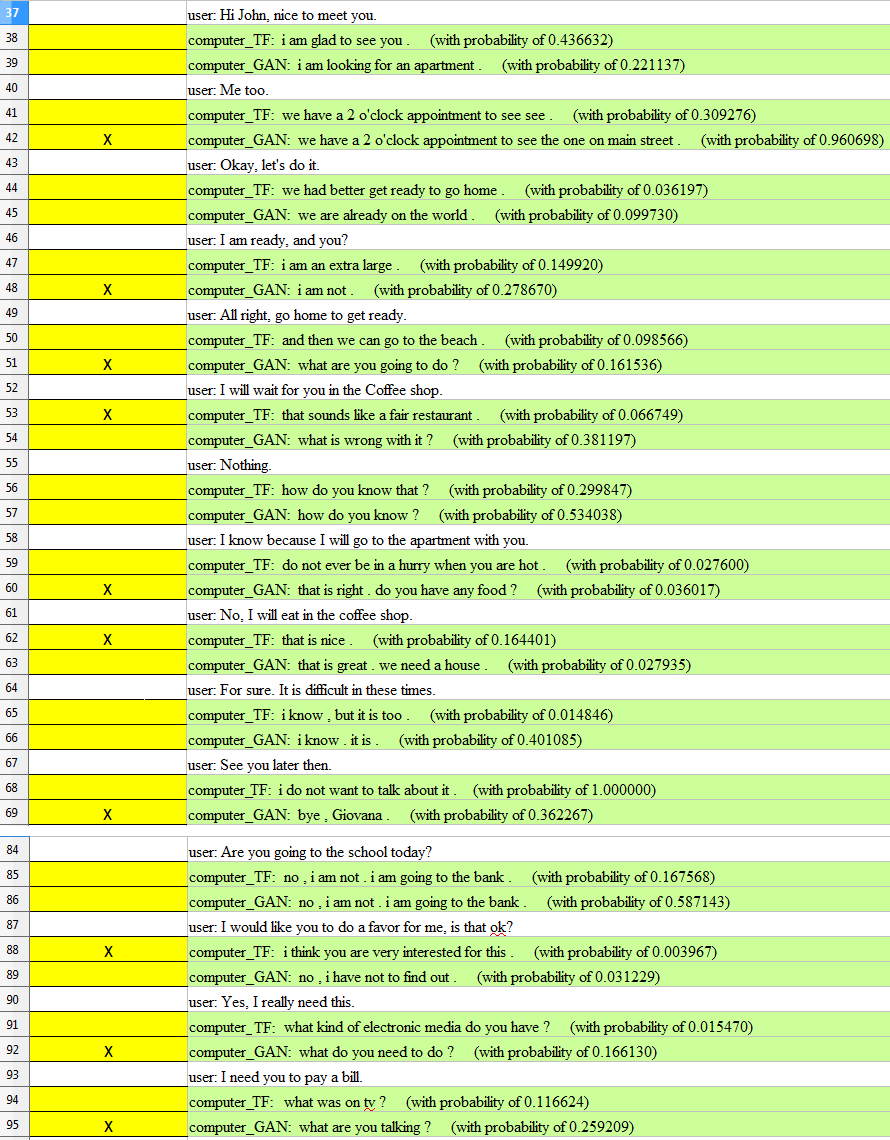}}
\caption{Samples from the evaluation form that has human-machine dialogue lines and yellow fields where the evaluator marks the best machine-generated answer.}
\label{eval_diag}
\end{center}
\vskip -0.3in
\end{figure}

The trained models chatted with humans to collect dialogue utterances for the test data set. Having this data set, we conducted a human survey with $25$ volunteers. Each respondent was given $10$ short dialogues, with an average of $24$ lines per dialogue, as can be seen in Figure \ref{eval_diag}.
\begin{flushleft}
\end{flushleft}

For each human-generated line corresponds a pair of answers, one from the model trained by teacher forcing and another from the model trained by the new adversarial method. The respondents can vote for the best machine-generated answer or assign a tie (no votes) per each dialogue line. Therefore, each machine-generated dialogue line receive a number of votes $v \in \mathbb{Z}: 0\leq v \leq 25$.

The discriminator $D$ is also used to have an adversarial evaluation on the machine-generated answers. The direct application of (\ref{prob}) favors short answers; therefore, we also adopt:
\begin{equation}
\label{prob2}
S\left(\textbf{y}^+\right) = \sqrt[s_s^*]{\prod_{i=1}^{s_s^*}l_i}
\end{equation}  
as a score to compare machine-generated answers, where $l_i$ is the discriminator output at the iteration $i$.

To determine the correlation between the human judgment and the score assigned by the discriminator, we computed the Jaccard index to measure the similarity between two sets of winning answers, one selected by the humans, $H$, and another by the adversarial method, $A$:
\begin{equation}
\label{Jaccard}
J(H,A) = \frac{|H \cap A|}{|H \cup A|}
\end{equation}
where $|H \cup A|$ is the total number of dialogue lines that received votes (remembering that in the event of a tie, there is no vote). The direct application of (\ref{prob}) yielded $J\left(H,A\right) = 0.41$, while using (\ref{prob2}) we obtained $J\left(H,A\right) = 0.58$. Tables \ref{evaluations1} and \ref{evaluations2} summarize the results using two different criteria: the best answer and the number of human votes, respectively.
 
\begin{table}[!hbt]
\renewcommand{\arraystretch}{1.3}
\caption{The number of best machine-generated answers. }
\label{evaluations1}
\centering
\footnotesize
\begin{tabular}{lccc}
\hline
training method & human evaluation & adversarial evaluation using (\ref{prob2}) \\
\hline
teacher forcing & 26  (29.88$\%$) & 35  (38.04$\%$) \\
adversarial learning & 61  (70.11$\%$) & 57  (61.96$\%$)\\
\hline
\end{tabular}
\end{table}

\begin{table}[!hbt]
\renewcommand{\arraystretch}{1.3}
\caption{Number of human votes}
\label{evaluations2}
\centering
\footnotesize
\begin{tabular}{lccc}
\hline
training method & human votes \\
\hline
teacher forcing & 252  (30.66$\%$) \\
adversarial learning & 570  (69.34$\%$) \\
\hline
\end{tabular}
\end{table}

If the absolute value of the difference between the scores is smaller than $5\%$ of the smallest score, the discriminator assigns a tie.

As can be seen in Tables \ref{evaluations1} and \ref{evaluations2}, the adversarial training yields a large gain in performance\footnote{The codes of our new adversarial training method are available at https://github.com/oswaldoludwig/Adversarial-Learning-for-Generative-Conversational-Agents }.

Even better performance is achieved by using the trained discriminator $D$ to choose the best answer among those generated by the two trained models. Our algorithm applies (\ref{prob2}) to rank the answers. An implementation of this method was also made available at GitHub\footnote{See the file conversation$\_$discriminator.py in the repository https://github.com/oswaldoludwig/Seq2seq-Chatbot-for-Keras. Follow the instructions in the README file to chat with the trained model.}.

The number of dialogue utterances used to compose the context $\textbf{x}$ is an important hyper-parameter. Our experiments indicate that with $N_u = 1$ the GCA cannot retain the context properly. On the other hand, with $N_u > 2$ the machine cannot change the subject of the conversation when the human interlocutor does it. An algorithm able to process the conversation and detect changes in the context would be useful to set the best value of $N_u$ before generating an answer.

\section{Conclusion}
\label{conclusion}

This work introduces a new model of GCA and a new adversarial training method for this model. The results presented in this paper indicate that the new training method can provide significant gains in performance. Moreover, the adversarial training also yields a trained discriminator that can be used to select the best answer, when different models are available.

As future work, we intend to evaluate the performance of hierarchical LSTMs in the encoder of our model and use an auxiliary algorithm to detect changes in the dialogue context, aiming at dynamically determining the optimal length of dialogue history to be encoded within the thought vector. It is also important to scale up the experiments by using larger data sets and evaluating larger versions of our model.

\def\bibsection{\section*{References}}



\bibliographystyle{elsarticle-num}

\bibliography{bib_file}

\end{document}